\documentclass[10pt,twocolumn,letterpaper]{article}

\usepackage{cvpr}
\usepackage{times}
\usepackage{epsfig}
\usepackage{graphicx}
\usepackage{amsmath}
\usepackage{amssymb}
\usepackage{algorithm}
\usepackage{algorithmic}

\usepackage[pagebackref=true,breaklinks=true,letterpaper=true,colorlinks,bookmarks=false]{hyperref}

\def\cvprPaperID{2223} 

\ifcvprfinal\fi
\begin{document}

\title{NegCut: Automatic Image Segmentation based on MRF-MAP}

\author{Qiyang Zhao\\
NLSDE, Beihang University\\
zhaoqy@buaa.edu.cn
}

\maketitle

\begin{abstract}
Solving the Maximum a Posteriori on Markov Random Field, MRF-MAP,
is a prevailing method in recent interactive image segmentation
tools. Although mathematically explicit in its computational
targets, and impressive for the segmentation quality, MRF-MAP is
hard to accomplish without the interactive information from users.
So it is rarely adopted in the automatic style up to today. In
this paper, we present an automatic image segmentation algorithm,
NegCut, based on the approximation to MRF-MAP. First we prove
MRF-MAP is NP-hard when the probabilistic models are unknown, and
then present an approximation function in the form of minimum cuts
on graphs with negative weights. Finally, the binary segmentation
is taken from the largest eigenvector of the target matrix, with a
tuned version of the Lanczos eigensolver. It is shown competitive
at the segmentation quality in our experiments.
\end{abstract}

\section*{1 Introduction}

Image segmentation is an important research field in computer
vision and computational graphics. In recent years, many
interactive segmentation methods, such as \textit{GrabCut}([1]),
\textit{Paint}([2]), are designed in the \textit{Bayesian}
framework, to say, \textit{Maximum a Posteriori} (MAP) on
\textit{Markov Random Fields} (MAP)([3]). Due to the high quality
of their segmentation results, especially of accurate boundaries
of extracted objects/foregrounds, these methods are capable of
processing many tedious and time-resuming interactive tasks in
image editing, medical diagnosing, etc.

In MRF-MAP-based segmentation methods, the sample foreground
and/or background zones are designated by user interactions first,
either in the scribble-based or painting-based style. The sample
data is then adopted to typically produce two appropriate
probabilistic models for the color distributions in the foreground
and background zones. Accompanied with some well-designed
smoothness terms, the data terms appear in the form of the
log-hoods of all probabilities of pixel colors, then compose the
final target function. The target function is then to be optimized
to obtain a good segmentation result. Clearly, the user
interactions are important in the whole task, in other word, it's
the first power to push the optimization to run. Especially when
the smoothness term only reflects the noisy tolerance capability,
as in \textit{GrabCut} and \textit{Paint}, the user-defined color
distributions are much critical in determining which part the
pixels belong to.

Compared with the generally dissatisfactory segmentation quality
in current automatic segmentation methods, it's tempting of
considering the good segmentation results in MRF-MAP-based
interactive tools. So an interesting problem arises naturally
here: whether or not we can design a segmentation algorithm in the
MRF-MAP framework, but without any user interactions? However
there are few concrete attempts except SE-Cut, to the best of our
knowledge. In order to produce the required color models, the
user-designated zones in \textit{GrabCut} are replaced by
automatically chosen seeds in SE-Cut([4]). SE-Cut chooses its
seeds by random walks and spectral embedding techniques instead in
the MRF-MAP framework itself. Clearly, it's not a unified
processing in only one dominating thread in the theoretical sense.
In practice, there's an additional but necessary requirement for
all seeds to be large enough lest being eliminated in the advanced
processing. However, it's hard to derive from the MRF-MAP
directly, and it makes SE-Cut a little more inconvenient both in
practice and theoretical views.

In this paper, we present an automatic segmentation algorithm,
NegCut, thoroughly under the MRF-MAP framework. It is designed to
produce binary segmentation results, and likely to figure the
dominating objects or outstanding scenes out of the images, the
same as the interactive segmentation tools. Our main contributions
are: (1) a mathematically strict analysis on the hardness of
MRF-MAP without any predefined information (or uninformed MRF-MAP
as abbr.) (2) a feasible way to approximate the MRF-MAP functions,
and a concise mapping from the optimization of approximating
energy onto minimum cuts of graphs with negative weights. (3) a
simple implementation with the help of the \textit{Lanczos}
eigensolver([5]), along with some experiments and corresponding
analysis.

\section*{2 Background}

In a MRF-MAP-based interactive segmentation algorithm, the task is
mapped into the minimization of a particular target energy
function $E$, which is the sum of a data term $E_{D}$ and a
smoothness term $E_{S}$ multiplied by a factor $\lambda$:
\begin{equation}
E=E_{D}+\lambda E_{S}
\end{equation}
where the data term $E_{D}$ is the sum of all data penalty on all
pixels according to their labels, the smoothness term $E_{S}$ is
the sum of all potential of adjacency interactions between all
neighboring pixels of different labels. In our situation, these
labels are just \textit{foreground} or \textit{background}
(\textit{fore} and \textit{back} as abbr.) Typically in current
algorithms, the data penalty of each pixel is the negative
log-hood of its probability according to the color distribution in
the foreground or background zone, and the smoothness terms are
usually defined on all adjacent pixels on 4/8-connection grids:
\begin{equation}
E_{D}=\sum\limits_p{-\ln Pr_{L(p)}(p)},E_{S}=\sum\limits_{(p,q)\in
N, L(p)\ne L(q)}{S(p,q)}
\end{equation}
where $L(p)$ is the label of pixel $p$ in the label configuration
$L$, and $(p,q)\in N$ means $p$ is adjacent to $q$. Although the
smoothness terms are critical and important also, it's not the
point in this paper. In general the smoothness term could be
regarded as a prior in form of a \textit{Markov Random Field},
therefore the problem is actually to pursue an appropriate label
configuration to reach the lowest energy $E$, which is the
\textit{Maximum a posterior} under the MRF-MAP framework. Although
it's more exact to regard the prior as a conditional random field
or discriminating random field, we here still take MRF as the
notation following the past literatures for convenience.

There are many choices when determining the coefficients in the
energy function $E$. For the data term, the ordinary statistical
histograms are sufficient for the situations with only a few
possible colors, such as 16 gray levels, 256 colors and so on.
However, it is not suitable for large color spaces such as 24-bit
true colors, as the samples were relatively too sparse to raise a
meaningful probabilistic model. Then other more complicated
models, such as the \textit{Gaussian mixture model }(GMM), fit
this situation better considering the compromise both of
efficiency and accuracy. It should be pointed out that, the GMM
method is compatible to the histogram method, if we take each
color value as an exact \textit{Gaussian} component with variance
0.

For the smoothness term, the coefficients should be designed to
encourage continuity between adjacent pixels, especially those of
similar colors. A good choice is the exponential function of the
\textit{Euclidean} distances of colors, as already adopted in
\textit{GrabCut}. Also we could add a posisstive constant to all
smoothness coefficients to enhance the local continuity, or reduce
the factor \textit{$\lambda $ }down to eliminate the differences
between different colors.

After the color probabilistic models and smoothness coefficients
are determined, there are several methods to minimize the target
energy function, such as graph cut, \textit{loopy belief
propagation} (LBP) and \textit{iterated conditional mode}
(ICM)([6]). In the case of binary labels, the graph cut method is
shown to be practically efficient: the target energy function
could be mapped into cuts on an undirected graph where the global
minimum solution could be obtained with the max-flow algorithms in
the polynomial time. Here the data penalties are regarded as the
weights of edges linked to the source/sink node in networks. As to
the case of multiple labels, however, the hardness changes, to
say, it's a k-way cut problem which is proven NP-hard.

The determined coefficients of the data terms, derived from the
probabilistic color models, are important to construct the graph
to be cut. Undoubtedly the user interactions are necessary for the
current MRF-MAP based interactive image segmentations, whereas
other tools such as random walks could present some candidate
foreground/background zones instead.

\section*{3 Hardness of uninformed MRF-MAP}

Remind that in the interactive image segmentations based on
MRF-MAP, the probabilistic color models are computed from sample
pixel colors, to be more detailed, foreground/background zones
designated by users. It means that the whole energy function is
still computable even if we only know the exact labels of all
pixels. From this point of view, now the energy function $E$ can
be written as

\begin{equation}
E(L)=\sum\limits_p {-\ln Gen(L,p)+\lambda \sum\limits_{(p,q)\in N,
L(p)\ne L(q)} {S(p,q)} }
\end{equation}
where we aim to obtain an appropriate probabilistic color model
from all pixel $p$'s with the same label $L_{p}$ in
{\{\textit{fore, back}\}}, with the help of a histogram- or
GMM-generating function \textit{Gen}. Given \textit{Gen}, all we
need to calculate $E$ is just the labels of all pixels in the
image. In other word, there is a corresponding energy value for
each label configuration of pixels. Clearly there exists a minimum
energy $E^{\ast}$ according to a certain label configuration
$L^{\ast}$ among them:
\begin{equation}
L^\ast =\mathop {\arg}\limits_L \min (E(L))
\end{equation}
Recall that we usually obtain a good segmentation result according
to the minimum energy value in the interactive segmentations, then
whether or not $L^{\ast}$ is also meaningful in the image
segmentation problem? Furthermore, could we obtain such a global
optimal solution $L^{\ast}$ in the polynomial time? However, it is
not addressed before to the best of our knowledge.

Since the histograms are easy to deal with, it's reasonable to
investigate the hardness of uninformed MRF-MAP on small number of
colors, while the conclusion still holds under the general cases
of GMMs. To be more concentrated, we neglect all smoothness term,
i.e., $\lambda $ = 0. Suppose we have $n$ pixels of $m$ colors in
the input image, and the pixels of color $i$ counts to $n_{i}$.
Then for each label configuration $L$ in {\{}\textit{fore},
\textit{back}{\}}$^{n}$, we set $n_{0, i}$ as the pixel amount of
color $i$ and label \textit{fore}, while it's the same for
$n_{1,i}$ to color $i$ and \textit{back}. Furthermore, $F$ and $B$
are the two sets of \textit{fore} pixels and \textit{back} pixels
respectively, while their pixel amounts are
$s_{0}=n_{0,1}+n_{0,2}+\ldots+n_{0,m}$ and
$s_{1}=n_{1,1}+n_{1,2}+\ldots+n_{1,m}$. These notations will also
be taken in the following sections. Now, the energy function $E$
on $L$ becomes
\begin{equation}
\begin{split}
 E(L)&=\sum\limits_{p\in F} {-\ln \frac{n_{0,p} }{s_0 }} +\sum\limits_{p\in
F} {-\ln \frac{n_{1,p} }{s_1 }} \\
 &=(s_0 \ln s_0+s_1 \ln s_1)-\sum\limits_{i=1}^m ({n_{0,i} \ln
 n_{0,i} + n_{1,i} \ln n_{1,i}})
\end{split}
\end{equation}

Then consider a function $f_{1}$ defined on interval $[0, c]$:
\begin{equation}
\begin{split}
f_1 (0)= &a\ln a+(b+c)\ln (b+c)-c\ln c,\\
f_1 (c)= &(a+c)\ln(a+c)+b\ln b-c\ln c,\\
f_1 (x) = &(a+x) \ln (a+x)+(b+c-x) \ln (b+c-x)\\
&- x \ln x - (c-x) \ln (c-x), (0<x<c).
\end{split}
\end{equation}
Obviously $f_{1}$ is continuous on the entire interval, and we
have its derivative ${f}'_1 (x)=0$ on $x=\frac{ac}{a+b}$, ${f}'_1
(x)>0$ when $0<x<\frac{ac}{a+b}$, and ${f}'_1 (x)<0$ for
$\frac{ac}{a+b}<x<c$. Hence the maximum value of $f_{1}$ on [0,
$c$] appears on $x=\frac{ac}{a+b}$, whereas its minimum value is
on $x=0$ or $x=c$. Then for each $k$ = 1, {\ldots}, $m$, if
$s_{0}$, $s_{1}$, $n_{k}$ and all $n_{0,i}$, $n_{1,i }(i\ne k)$ as
fixed, we would get the minimum value of $E(L)$ at $n_{0,k }$= 0
or $n_{1,k }$= 0. Without the loss of generality, we assume there
exists an $m^{\ast }$ satisfying that $n_{1,k }$= 0 for all $k \le
m^{\ast }$, and $n_{0,k }$= 0 for all $k > m^{\ast }$. So we have
\begin{equation}
\begin{split}
minE(L) = &(\sum\limits_{k\le m^\ast } {n_k } )\ln
(\sum\limits_{k\le m^\ast } {n_k } )+(\sum\limits_{k>m^\ast } {n_k
} )\ln
(\sum\limits_{k>m^\ast } {n_k } )\\
&-\sum\limits_{k=1}^m {n_k \ln n_k }
\end{split}
\end{equation}
Now consider another function $f_2 (x)=x\ln x+(d-x)\ln (d-x)$.
Since${f}'_2 (x)=\ln \frac{x}{d-x}$, it's easy to conclude that
$f_{2}$ reaches its minimum value at $x=\frac{d}{2}$. So finally
we know that the energy function $E$ reaches its minimum value
when$\sum\nolimits_{k\le n^\ast } {n_k } =\sum\nolimits_{k>n^\ast
} {n_k } $ if possible. In the following we will reduce the set
partition problem to the minimization of our energy function $E$.
Since the set partition problem is an NPC problem([7]), uninformed
MRF-MAP is NP-hard then. Remember that, the set partition problem
is about

\textit{Given a positive integer set} $X=${\{}$x_{1}$, $x_{2}$,
{\ldots}, $x_{m}${\}}\textit{ summing up to 2k, whether or not
there is a subset X}${'}$\textit{ satisfying that the sum of all
entries in X}${'}$\textit{ equals k?}

The reduction is rather straightforward. First we construct an
image \textit{img} with the size $1\times 2k$ (only one pixel
high) in the color space {\{}\textit{color}$_{1}$,
\textit{color}$_{2}$, {\ldots}, \textit{color}$_{m}${\}} as

\begin{figure}
\centerline{\includegraphics[width=3.5in,height=1.0in]{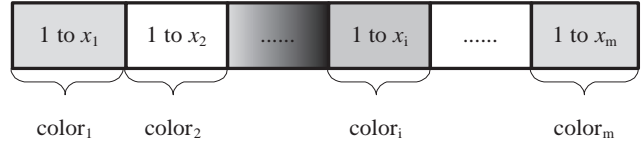}}
\caption{An $1\times 2k$ image with $m$ colors.}
\end{figure}

Then in the smoothness terms, we make it sufficiently large for
pixels of the same colors, but close to zero for pixels of
different colors. Therefore the blocks of identical colors would
never be cut, and the extremely small coefficients for adjacent
pixels of different colors would have no effect on the energy
function $E$. When minimizing $E$, we would get
\begin{equation}
min\left( {\left( {\sum\limits_{i\in {X}'} {x_i } } \right)\ln
\left( {\sum\limits_{i\in {X}'} {x_i } } \right)+\left(
{\sum\limits_{i\in \bar {{X}'}} {x_i } } \right)\ln \left(
{\sum\limits_{i\in \bar {{X}'}} {x_i } } \right)} \right)
\end{equation}
on a certain subset $X{'}$ satisfying $\sum\nolimits_{i\in {X}'}
{x_i } =\sum\nolimits_{i\in \bar {{X}'}} {x_i } =k$, according to
the above analysis. So the set partition problem could be
determined by minimizing the energy function $E$ and verifying
whether the minimum is equal to $\left( {2k\ln
k-\sum\nolimits_{i=1}^m {x_i \ln x_i } } \right)$. So the
reduction is finished.

Here we have established the link between the set partition
problem and the uninformed MRF-MAP in image segmentations. Recall
that, the hardness of normalized cut is also linked to the same
NPC problem([8]), and the reduction there implicitly ensures that
its binary segmentation results are somewhat fair in sizes.
Clearly this favorable property also holds for the potential image
segmentation algorithms based on uninformed MRF-MAP, to say, no
much isolated small-sized pieces were produced.

\section*{4 Approximating uninformed MRF-MAP}
Considering the NP-hardness when minimizing the energy function
$E$, it is reasonable to pursue its approximating solution
instead. However, $E$ is not a polynomial with respect of the
label configuration $L$, even not a closed form at all. So it is
feasible to find its replacement which is solvable and
sufficiently close to $E$.

First let's consider a function$f_3 (x)=x\ln x+(1-x)\ln (1-x)$. We
have $f_3 (x)=\frac{-5}{2}xy-\Delta (x)$ where $y$ = (1 -- $x)$
and
\begin{equation}
\Delta (x)=xy(\frac{1}{3}(x^2+y^2)+\frac{1}{4}(x^3+y^3)+\cdots ),
\end{equation}
by performing two \textit{Taylor} expansions. The mean of
\textit{$\Delta $}($x)$ on [0, 1] is $\int_0^1 {\Delta
(x)dx=\frac{1}{12}} $, and we could basically taken it as the
value of \textit{$\Delta $}($x)$ on the entire interval, given
that the mean square error $\int_0^1 {(\Delta
(x)-\frac{1}{12})^2dx\approx }\ 3\times 10^{-4}$ is considerable
small. Hence it is totally acceptable to approximate $f_{3 }$ with
$\left( {\frac{-5}{2}x(1-x)-\frac{1}{12}} \right)$, as shown in
Fig. 2.
\begin{figure}[htbp]
\centerline{\includegraphics[width=3.1in,height=2.0in]{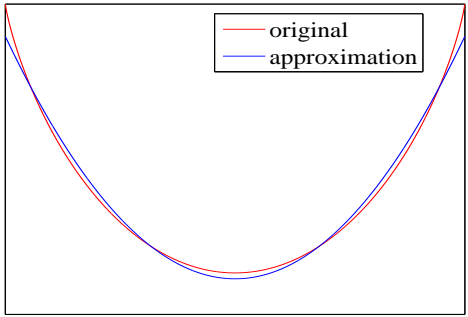}}
\caption{Curves of function $f_3$ and its approximation on [0,
1].}
\end{figure}
Now return to our energy function $E$. Since the uninformed
MRF-MAP on small number of colors is relatively simple, we set the
case as our start.

\subsection*{4.1 For small color spaces}
Histograms are generally adopted in the cases of small color
amounts. Based the approximation of $f_{3}$, we can approximate
the energy function $E$ with
\begin{equation}
\begin{split}
( {\frac{-5}{2n}s_0(n-s_0 )-\frac{1}{12}} )&-\sum\limits_{i=1}^m
{\left( {\frac{-5}{2n_i }n_{0,i} (n_i
-n_{0,i} )-\frac{1}{12}} \right)} \\
&+\lambda \sum\limits_{(p,q)\in N, L(p)\ne L(q)} {S(p,q)}
\end{split}
\end{equation}
Since $m$, $n$ and all $n_{i}$'s are all fixed in the input image,
it is actually to minimize
\begin{equation}
\begin{split}
\frac{-5}{2n}s_0 (n-s_0 ) &-\sum\limits_{i=1}^m {({\frac{-5}{2n_i }n_{0,i} (n_i-n_{0,i} )} )}\\
&+\lambda\sum\limits_{(p,q)\in N,L(p)\ne L(q)} {S(p,q)}
\end{split}
\end{equation}
when pursuing the minimum of $E$.

\subsection*{4.2 For large color spaces}
It is a little more complicated for the cases of large color
amounts, such as $2^{24}$ colors in the 3-bytes RGB space. Since
the amount of potential colors is usually much larger than that of
the image pixels, there is hardly remarkable number of samples in
each bin of the color histogram. \textit{GrabCut} takes GMM as a
good replacement to histograms in the color image segmentation,
but it is not fluent and easy to find the approximation of $E$
from GMMs. Instead we adopt a trivial but also accurate scheme in
the continuous RGB color space: first, choose a fidelity parameter
$\sigma ^{2}$ empirically or by calculations on the entire image
as in [1], then set the statistical contribution of sample color
on each pixel $p$ to be a normal distribution with mean $p$ and
variance $\sigma ^{2}$
\begin{equation}
Pr_{p}(k)=\frac{1}{\sqrt {2\pi \sigma ^2}
}e^{-\frac{Dist(p,k)}{2\sigma ^2}}
\end{equation}
where $Dist(p,k)$ means the squared \textit{Euclidean} distance
between the colors of pixel $p$ and $k$. Then construct the
probabilistic color models on all foreground pixels and background
pixels as
\begin{equation}
Pr_{F} (p)=\frac{1}{n_0 }\sum\limits_{q\in F} {Pr_{q}(p)} ,Pr_{B}
(p)=\frac{1}{n_1 }\sum\limits_{q\in B} {Pr_{q}(p)}
\end{equation}
And the energy function $E$ becomes into the smoothness term plus
an integral on the entire continuous RGB space
\begin{equation}
\begin{split}
E(L)=&\int \left(-{\frac{n_0 }{n}} \cdot {Pr_{F} (p)\ln Pr_{F}(p)} \right)dp\\
&+\int \left(-{\frac{n_1 }{n}} \cdot {Pr_{B}(p)\ln Pr_{B} (p)} \right)dp\\
&+\lambda \sum\limits_{(p,q)\in N,L(p)\ne L(q)} {S(p,q)}
 \end{split}
\end{equation}
Since$\sum\nolimits_q {\Pr _q (p)} $ is a fixed value in the input
image, it is actually to minimize
\begin{equation}
\begin{split}
&\sum\limits_{p,q} {\left( {-\frac{5}{2n^2}+\frac{5}{2n}\cdot \int
{\frac{\Pr _{p} (k)\cdot \Pr _{q} (k)}{\sum\nolimits_j
{\Pr _j (k)} }dk} } \right)} \\
&+\lambda \sum\limits_{(p,q)\in N,L(p)\ne L(q)} {S(p,q)}
 \end{split}
\end{equation}
when pursuing the minimum of $E$. Again we can make it simpler by
enlarging the factor \textit{$\lambda $ } $n$ times
\begin{equation}
\begin{split}
&\sum\limits_{p,q} {\left( {-\frac{5}{2n}+\frac{5}{2}\cdot \int
{\frac{\Pr _{p } (k)\cdot \Pr _{q } (k)}{\sum\nolimits_j
{\Pr _j (k)} }dk} } \right)} \\
&+\lambda \sum\limits_{(p,q)\in N,L(p)\ne L(q)} {S(p,q)}
 \end{split}
\end{equation}
\section*{5 NegCut based on Min-Cut}

\begin{figure*}
\centerline{\includegraphics[width=6.0in,height=2.3in]{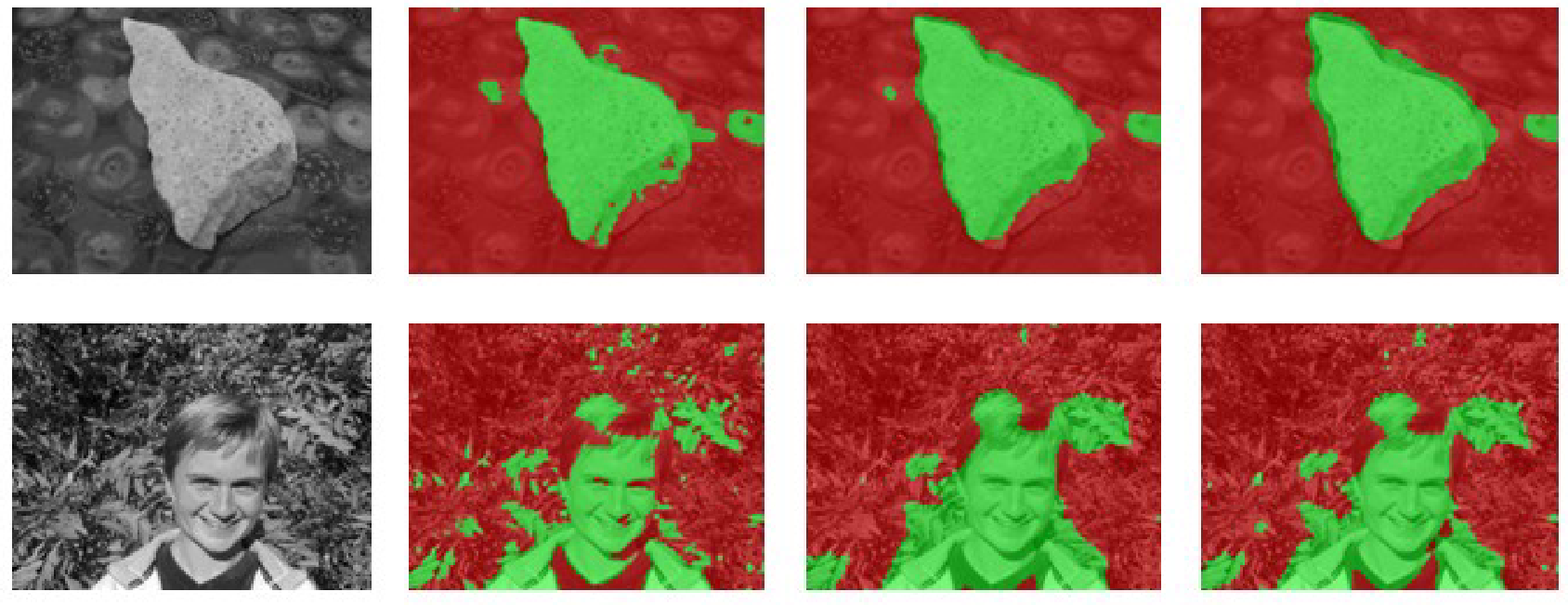}}
\caption{Leftmost column: original gray images, other three
columns: segmentation results with $\lambda=1, 5, 10$ in turn.
($16 gray scales$)}
\end{figure*}

\subsection*{5.1 NegCut on small color spaces}
Take the simple case of histograms on small number of colors as
our start. Then our target is to obtain a segmentation result by
minimizing (11). Now construct an undirected graph $G $ of $n$
nodes corresponding to the pixels, and set the edge weight $w(p$,
$q)$ as the sum of the following three terms
\begin{equation}
\begin{split}
&w_1 (p,q)=\frac{-5}{2n} \\
&w_2 (p,q)=\left\{ {\begin{array}{l}
 \frac{5}{2n_i },p\mbox{ and }q\mbox{ have the same color }i, \\
 \mbox{0,otherwise.} \\
 \end{array}} \right.\\
&w_3 (p,q)=\left\{ {\begin{array}{l}
 S(p,q),p\mbox{ is adjacent to }q, \\
 \mbox{0,otherwise.} \\
 \end{array}} \right.
\end{split}
\end{equation}
It is easy to prove that the approximating energy value on a label
configuration $L$ is just equal to the capacity of the
corresponding cut $C $= {\{}$F$, $B${\}} on $G$. Therefore it is
to find the minimum cut on $G$ minimizing (11). Although there
exists several O($N^{3})$ algorithm solving the minimum cuts on
graphs with non-negative weights, it is not suitable here. In
fact, the negative weighted edges make the Min-Cut to be a Max-Cut
on a auxiliary graph with non-negative weights. Since Max-Cut has
already be proven to be a NP-complete problem, it can be concluded
that there is no polynomial algorithm for us to minimize (11).

However, we might generalize this discrete optimization task into
the continuous real space $R^n$. First we set an indicator vector
$D=\left[ {d_1 ,d_2 ,\cdots ,d_n } \right]^T\in \left\{ {+1,-1}
\right\}^n$ denoting the $i$th pixel of \textit{fore} by $+1$ or
\textit{back} by $-1$, then establish a matrix $W=[w(p,q)]$ whose
entries sum up to $S_{W}$. Clearly the cut value is just equal to
$\frac{1}{2}\left( {S_W -D^TWD} \right)$. Subsequently, we
generalize the labels into the continuous interval $[-1, 1]$
instead of $ \{+1,-1\} $, and the original optimization task on
(11) becomes
\begin{equation}
\max D^TWD,\mbox{s.t.}\left\| D \right\|_2 =n.
\end{equation}
According to the \textit{Lagrange} \textit{factor}
\textit{method}([6]), the solution to (18) is the eigenvector $D$
corresponding to the largest eigenvalue. Then we might
straightforwardly get the required binary indicator from $D$ by
setting the $i$th entry as $-1$ if $d_{i} \le 0$, or $+1$ if
$d_{i} > 0$. Here the \textit{Lanczos} algorithm is adopted to
finish the time-consuming operations of the largest eigenvectors
for large matrices. In fact, the \textit{Lanczos} algorithm is
well known as the fastest method to solve the largest eigenvectors
in $O(N)$ time mostly for sparse matrices, whereas our graph $G$
is completely connected and the corresponding weight matrix $W$ is
full. Fortunately, $W$ is only called to perform a matrix-vector
multiplication there, and the major calculation is identical for
pixels of the same color. So we might improve it to sharply reduce
the time consumption down to linear, as described in algorithm 1.

Now we have established the NegCut algorithm on the small color
spaces as below:

\begin{algorithm}
\caption{NegCut on small color spaces}
\begin{algorithmic}[1]
\STATE On each adjacent pair of pixels ($p$, $q)$, compute the
smoothness term $S(p$, $q)$;

\STATE Solve the largest eigenvector $D=\left[ {d_1 ,d_2 ,\cdots
,d_n } \right]^T$ of the matrix $W=\left[ {w(p,q)} \right]$ with
the \textit{Lanczos} algorithm, but perform the matrix-vector
multiplication of $W$ and some vector $R=\left[ {r_1 ,r_2 ,\cdots
,r_n } \right]^T$ as

\begin{itemize}
\item set \textit{$\varphi $} = 0. Then for $k$ = 1 to $n$, $\varphi
\leftarrow \varphi -\frac{5}{2n}\cdot r_k $;

\item set \textit{$\theta $}$_{i}$ = 0 for all $i$'s. Then for $k$ =
1 to $n$, set $\theta _i \leftarrow \theta _i +\frac{5}{2n_i
}\cdot r_k $ according to its color $i$;

\item for $k = 1$ to $n$, calculate $\mu _k
=\sum\nolimits_{j:(j,k)\in N} {S(j,k)} $. Then according to its
color $i$, and output $\varphi +\theta _i
+(\frac{5}{2n}-\frac{5}{2n_i })\cdot r_k $ as the $k$th entry of
vector $W \cdot R$.

\end{itemize}

\STATE let the label of pixel $k$ be \textit{back} if $d_{k} \le
0$, or \textit{fore} if $d_{k} > 0$.

\end{algorithmic}
\end{algorithm}

Apparently we could finish step 1 and 3 in $O(n)$ time. From step
2.a to 2.d, it requires $O(n)$ time to calculate the product
vector involving $W$, so we need totally $O(n)$ time for the
slightly changed \textit{Lanczos} algorithm in step 2. Finally the
entire time complexity of NegCut for small color spaces remains
$O(n)$.

\subsection*{5.2 NegCut on large color spaces}
The case of large color spaces is a little more complicated.
Recall that in (16) we need to calculate the integral on the whole
continuous color space, and it is unrealistic in most
applications. A feasible substitution is to perform the
calculation on the typical samples according to the probabilistic
distribution of (13). Fortunately the pixels of the input image
just meet this requirement well. Hence the integral becomes into
the sum as $\sum\nolimits_k {\frac{\Pr_{p} (k)\cdot \Pr_{q}
(k)}{\sum\nolimits_j {\Pr _j (k)} }} $, however the computational
load is still heavy as we have to scan each pixel and record its
contributions to $n$ Pr$_{q}$'s. While the entire complexity is up
to $O(n^{2})$, which is totally unacceptable even on
ordinary-sized images. In order to improve it, we cluster all
colors into a limited number of classes first, so that the
summations would be identical for the pixels in the same color
class. Now, all we need is just to scan each pixel and record its
contributions to these color classes, and the time consumption is
reduced to $O(n)$ given the amount of color classes is limited.

\begin{figure*}
\centerline{\includegraphics[width=5.5in,height=5.25in]{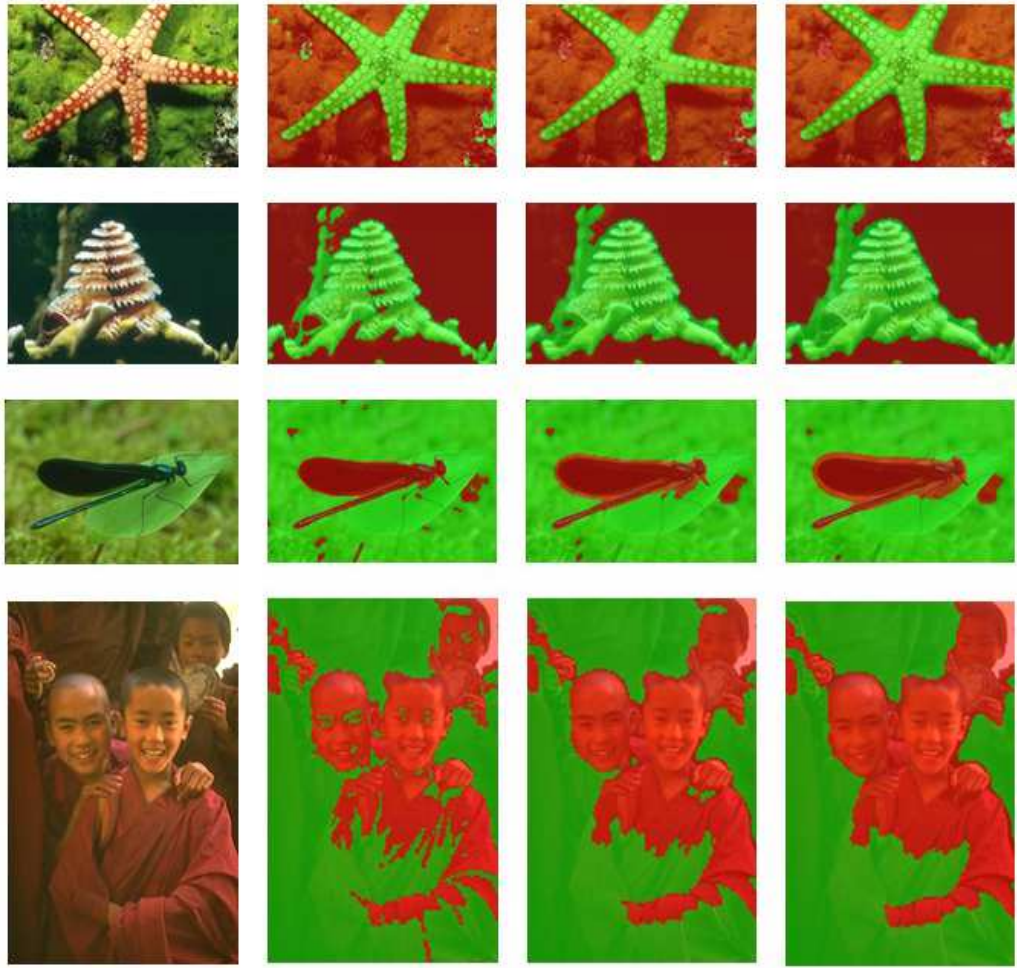}}
\caption{Leftmost column: original images, other three columns:
segmentation results with $\lambda=1, 5, 10$ in turn. ($m=16$)}
\end{figure*}

As in the small color spaces, we construct an undirected graph $G$
of $n$ nodes corresponding to the pixels, and set the edge weight
$w(p,q)$ as the sum of following three terms,
\begin{equation}
\begin{split}
&w_1 (p,q)=\frac{-5}{2n} \\
&w_2 (p,q)=\frac{5}{2}\cdot \sum\nolimits_k {\frac{\Pr _p (k)\cdot
\Pr _q (k)}{\sum\nolimits_j {\Pr _j (k)} }} \\
&w_3 (p,q)=\left\{ {\begin{array}{l}
 S(p,q),p\mbox{ is adjacent to }q, \\
 \mbox{0,otherwise.} \\
 \end{array}} \right.
 \end{split}
\end{equation}
Now we have established the NegCut algorithm for the large color
spaces:

\begin{algorithm}
\caption{NegCut on large color spaces}
\begin{algorithmic}[1]
\STATE Cluster all pixels into $m$ color classes with the mean
color $c_{1}$, {\ldots}, $c_{m}$. On each pixel $k$, calculate its
probabilistic densities in all $m$ Gaussian Distributions, and add
its contribution $\frac{\Pr _p (k)\cdot \Pr _q
(k)}{\sum\nolimits_j {\Pr _j (k)} }$ to $w_{2}(p$, $q)$ for every
two color classes of pixel $p$ and $q$;

\STATE Solve the largest eigenvector $D=\left[ {d_1 ,d_2 ,\cdots
,d_n } \right]^T$ of the matrix$W=\left[ {w(p,q)} \right]$ with
the \textit{Lanczos} algorithm, but perform the matrix-vector
multiplication of $W$ and some vector $R=\left[ {r_1 ,r_2 ,\cdots
,r_n } \right]^T$ as

\begin{itemize}
\item set \textit{$\varphi $} = 0. Then for $k = 1$ to $n$, $\varphi \leftarrow \varphi -\frac{5}{2n}\cdot r_k $;
\item set \textit{$\theta $}$_{i}$ = 0 for all $i$'s. Then for $k$ = 1 to $n$, $i$ = 1 to $m$, set $\theta _i \leftarrow \theta _i +w_2 (j,k)\cdot r_k
$ according to its color class $j$;
\item for $k=1$ to $n$, calculate $\mu_k =\sum\nolimits_{j:(j,k)\in N}{S(j,k)}$. Then according to its color class $i$, output $\varphi +\theta _i
+\mu _k -\frac{5}{2n}\cdot r_k $ as the $k$th entry of vector $W
\cdot R$.
\end{itemize}
\STATE let the label of pixel $k$ be \textit{back} if $d_{k} \le
0$, or \textit{fore} if $d_{k} > 0$.

\end{algorithmic}
\end{algorithm}

There are some light differences between the two algorithms in
step 1 and 2. However given the amount of color classes is limited
and fixed, it requires also O($n)$ time to finish the
multiplication involving $W$. Finally the entire time complexity
of NegCut for large color spaces is also O($n)$.

\section*{6 Experiments}
Since NegCut is different on the small color spaces and large
color spaces, our experiments are finished on both gray scale
images and color images. We trivially set all smoothness terms to
be 1 over entire images, and let $\lambda$ vary from 1 to 10. All
color test images are chosen from the segmentation datasets from
Berkeley and MSR at Cambridge, and all gray scale images are
derived from them. Here the amounts of color classes in all color
images are set to be 16, and all gray-scaled images also have 16
gray levels within them. Based on the block-based Lanczos
eigensolver([9]), we established two versions for NegCut in Matlab
codes. At last, all test images are resized to be about $256
\times 256$ to reduce the time-consuming calculations in the
experiments.

There are three groups of results in Fig. 3, 4 and 5: gray-scaled
on different $\lambda$'s, segmentation results of color images on
different $\lambda$'s, and different color images on fixed
$\lambda=1$. Our target images are intendly chosen to be of close
gray scales, splendidly colors and delicate local changes, so that
to verify the performance of NegCut on different challenges. The
small color space version of NegCut works on gray-scaled images,
while those color images are processed with the more complicated
algorithm.

In general, we obtained basically acceptable segmentation results
in all experiments, especially that most dominating objects are
figured out of these images. There are more isolated, but vivid
pieces in the segmentation results when $\lambda=1$, both on
gray-scaled images and color images. And it's much better for
color images because the connections between different colors are
involved in the large color space version of NegCut. When
$\lambda=1$, the segmentation boundaries are more likely located
on the desired edges of the objects in these images. However, it
also brings too much emphasis on these discontinuous line
segments, and results in much more isolated pieces in the
segmentations. When $\lambda$ varies from 1 to 5, then to 10, it
is shown the segmentation boundaries become smoother and smoother,
but on the cost of the lose of elaborate details. When
$\lambda=10$, the two segmentation zones are more fair in sizes.
The reason for that is, the large portion of the smoothness terms
in the energy function weakens the effect of the data penalties.
NegCut tends to cut two pieces sized of one half of the entire
image to reach the minimum energies.

\section*{7 Conclusions}
Despite the NP-hardness of uninformed MRF-MAP, NegCut is an
initial attempt in developing automatic image segmentation tools
under the MRF-MAP framework. Given that the Lanczos algorithm is
specially designed for solving the extreme eigenvectors, NegCut
are much easier to develop in principle than those require inner
eigenvectors, such as the normalized cut. Though NegCut only
produce binary segmenation results, it can be recurrently called
to refine the past segmentationzones results until the expected
one appears.

Recall that in the interactive segmentation tools, their target is
to obtain the minimum of particular energy functions with user
interactions. Since NegCut approximates the global minimum energy
value, it is meaningful in the analysis on the underlying
mathematical trends when these interactive tools performing
calculations.

Different from the ordinary minimum-cut-based segmentation or
clustering algorithms, the negative weights ensure that NegCut
likely obtain fair segmentation results instead of isolated
extremely small pieces. However, the double-fold NP-hardness
encountered in solving MRF-MAP reminds us, it is necessary to make
a attempt on more efficient calculation methods.

\begin{figure*}
\centerline{\includegraphics[width=4.8in,height=6.0in]{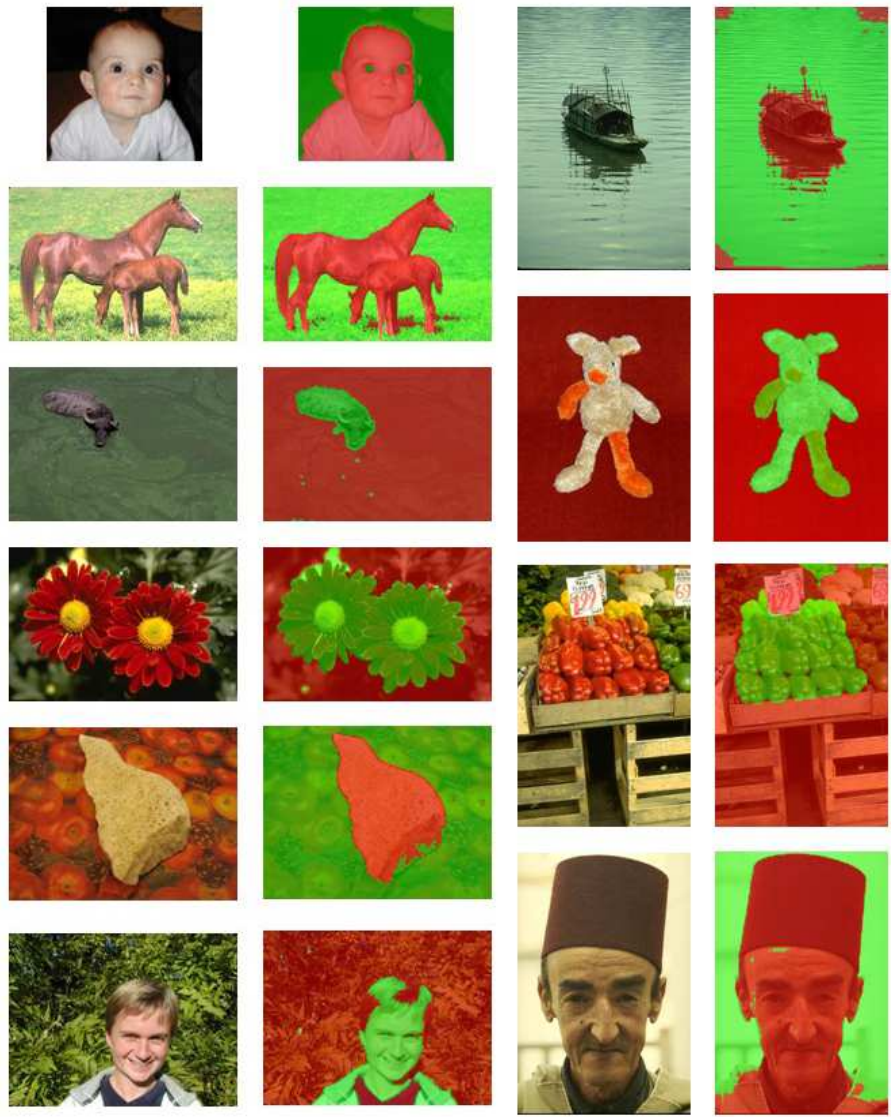}}
\caption{Left columns: original images, right columns:
segmentation results. ($\lambda=1, m=16$)}
\end{figure*}

{\small
\bibliographystyle{ieee}
\bibliography{egbib}
}

\end{document}